\pdfoutput=1

\documentclass[11pt]{article}

\usepackage[preprint]{acl}

\usepackage{times}
\usepackage{latexsym}

\usepackage[T1]{fontenc}

\usepackage[utf8]{inputenc}

\usepackage{microtype}
\usepackage{inconsolata}
\usepackage{graphicx}

\usepackage{placeins}
\usepackage{longtable}
\usepackage{acronym}
\usepackage{subfigure}
\usepackage{stfloats}
\usepackage{adjustbox}
\usepackage{breqn}
\usepackage{amsmath}
\usepackage{url}
\usepackage{comment}

\title{Generating Synthetic Free-text Medical Records with Low\\Re-identification Risk using Masked Language Modeling}

\author{Samuel Belkadi$^1$, Libo Ren$^2$, Nicolo Micheletti$^3$, \\ \textbf{Lifeng Han$^2$, Goran Nenadic$^2$} \\ 
$^1$ Department of Engineering, University of Cambridge, UK \\ 
$^2$ Department of Computer Science, The University of Manchester, UK \\ 
$^3$ 
Department of Computer Science and Technology, Tsinghua University, China \\
}

\begin{document}

\maketitle

\begin{abstract} 
The vast amount of available medical records has the potential to improve healthcare and biomedical research. However, privacy restrictions make these data accessible for internal use only. 
Recent works have addressed this problem by generating synthetic data using Causal Language Modeling. Unfortunately, by taking this approach, it is often impossible to guarantee patient privacy while offering the ability to control the diversity of generations without increasing the cost of generating such data.
In contrast, we present a system for generating synthetic free-text medical records using Masked Language Modeling. The system preserves critical medical information while introducing diversity in the generations and minimising re-identification risk. The system's size is $\sim120$M parameters, minimising inference cost. 
The results demonstrate high-quality synthetic data with a HIPAA-compliant PHI recall rate of 96\% and a re-identification risk of 3.5\%. Moreover, downstream evaluations show that the generated data can effectively train a model with performance comparable to real data.
\end{abstract}

\section{Introduction}
The adoption of electronic medical record systems has resulted in vast amounts of patient data with significant potential to enhance healthcare and biomedical research \cite{beam2018big, shah2018big}. However, privacy restrictions limit data accessibility to protect patients' private information \cite{price2019privacy}. Synthetic data provides a viable solution by generating records, such as discharge summaries, that maintain useful medical information with minimal privacy concerns. This can facilitate data sharing for applications such as health system testing \cite{tucker2020generating}, medical education \cite{li2023large}, and AI development \cite{belkadi2023exploring}.

Previous works on medical synthetic data generation have focused extensively on using Causal Language Modeling, while giving very little attention to Masked Language Modeling. 
Although the former demonstrates the ability to replicate the statistical properties of medical records, three main challenges are observed, namely the guarantee that privacy is not breached, the ability to control the diversity of generations, and the cost of generation.

Recent work by \newcite{micheletti2024exploration} shows that Masked Language Modeling (MLM) matches Causal Language Modeling (CLM) performance at most synthetic generation tasks, with greater control over the generations' context.
Supported by their discoveries, our paper introduces a system for generating English synthetic free-text medical reports, including discharge summaries, admission notes, and doctor correspondences, using Masked Language Modeling. 
The system incorporates a state-of-the-art de-identification tool for detecting protected health information \cite{radhakrishnan2023certified}, eliminating the need for prior manual de-identification. In addition, it uses two entity recognition models to preserve critical medical information and control the diversity-fidelity trade-off in generations. Finally, by using an encoder-only architecture that is not autoregressive, both the system's size and inference cost are significantly reduced.
The code will be publicly available.

\section{Related Work}
\label{sec:background}
In their recent work, \newcite{yan2024generating} introduced a Generative Adversarial Network for generating synthetic electronic health records. Their results showed limitations in controlling the resemblance between synthetic and original data, and the inability to capture temporal medical relationships.

Using similar methods, \newcite{kasthurirathne2021generative} developed a system to generate synthetic medical records with low re-identification rate. Although the results were promising, the authors claimed that the restricted diversity of the synthetic samples limited their applicability to tasks such as oversampling. Moreover, they assumed synthetic generation to inherently reduce re-identification risk, implying the need for further de-identification.

Finally, in one of the latest works on synthetic medical data, \newcite{falis2024can} evaluated GPT-3.5 at generating discharge summaries. Their results demonstrated that it often reproduced most concepts from prompts, increasing re-identification risk. Additionally, GPT-3.5 generated unnatural text, omitting critical information and introducing spurious content. Clinician evaluators noted ``correctness in generated summaries, but deficiencies in variety, supporting information, and narrative coherence''. Furthermore, the model may raise privacy concerns as it is not owned or controlled by the original data's custodian.

A clear pattern emerges between previous works on synthetic data generation. The main observations are that privacy often remains an issue and that the control over generations is usually limited. For these reasons, our work suggests that Masked Language Modeling can reduce privacy concerns and improve control over diversity, while minimising the cost of generating synthetic data.

\section{System Design}
\label{sec:design}

\begin{figure*}[t!]
  \centering
  \includegraphics[width=.88\textwidth]{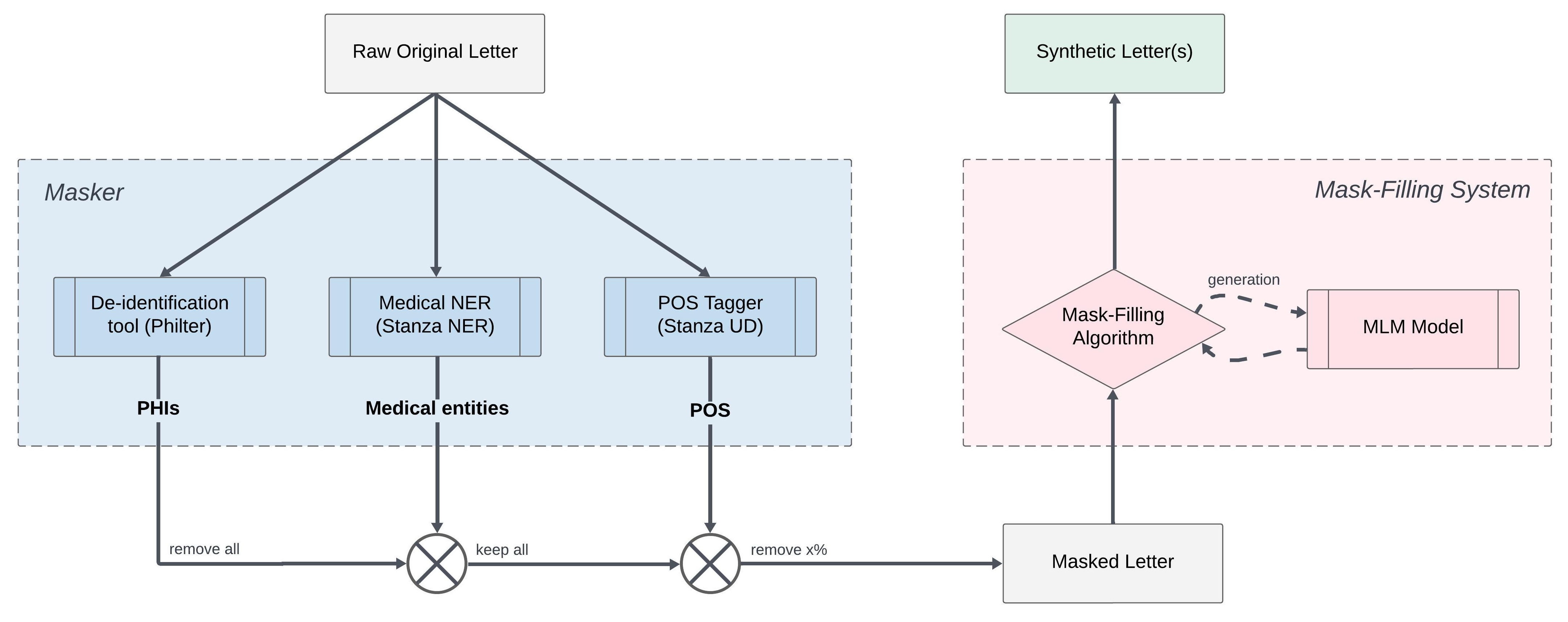}
  \caption{Design of the entire system, showcasing the Masker and Mask-Filling components.}
  \label{fig:system}
\end{figure*}

Our system displayed in Figure~\ref{fig:system} generates synthetic medical records, including discharge summaries, admission notes, and correspondences between doctors, through a two-step pipeline composed of a \textit{Masker} and a \textit{Mask-Filling System}. The Masker identifies entities to mask or retain, producing a masked letter as output. Subsequently, the Mask-Filling System replaces masked entities based on their context, generating one or more synthetic versions of the original letter.

\subsection{The Masker} 
The Masker operates in three consecutive phases: \\

\textbf{De-identification.} The first phase identifies Protected Health Information (PHI) using Philter \cite{norgeot2020protected}, a tool that employs regular expressions to extract six PHI categories (DATE, ID, NAME, CONTACT, AGE, LOCATION). The authors reported high recalls of 99.46\% on the UCSF dataset and 99.92\% on the i2b2 dataset of 2014. To the best of our knowledge, it is the first certified de-identification pipeline that makes clinical notes available to researchers for nonhuman subjects’ research without the need for further IRB approval, under the period specified by \citeauthor{radhakrishnan2023certified}. 

\textbf{Medical Entity Recognition.} The second phase uses a medical named entity recognition (NER) model to identify key medical entities to retain in the synthetic letter. We fine-tune a pre-trained instance of Stanza\footnote{stanfordnlp.github.io/stanza/available_biomed_models.html} on the i2b2-2010 dataset to extract three types of entities, namely PROBLEM, TEST, and TREATMENT, achieving an F1 score of 88.13\% on the testing data. Depending on the application, the model can be replaced to identify different entities (e.g., medications and dosages) and masking ratios can be adjusted to control how much of each category should be retained.

\textbf{Part-of-Speech Tagging.} The final phase uses Stanza's POS tagger to identify parts of speech in the remaining text. A subset of tagged entities is randomly masked based on user-defined ratios to further control the diversity in the synthetic outputs. For example, one could define the mapping \{NOUN: 0.7, VRB: 0.5\} to mask 70\% of nouns, 50\% of verbs and none of the other categories.

\subsection{The Mask-Filling System} 
Given the masked letters produced by the Masker, the Mask-Filling System uses an MLM model and a Mask-filling algorithm to generate synthetic letters. \\

\textbf{MLM Model.} The MLM model is an encoder model which provides a probability distribution over all possible words to replace the masked entities with respect to their context. The system employs Bio\_ClinicalBERT, an instance of BioBERT \cite{lee2020biobert} fine-tuned on clinical notes from MIMIC III \cite{johnson2016mimic}. We further train this model for our task on the 790 letters provided by the dataset described in Section~\ref{sec:dataset}. Please note that we did not try alternative baseline models. However, we truly encourage further studies to experiment with that. \\

\textbf{Mask-Filling Algorithm.} This component prepares chunks of masked text for the MLM model and selects replacements from the vocabulary based on the model's output probabilities. We compare two mask-filling approaches detailed below:

\begin{itemize}
    \item \textit{Simultaneous Chunk Filling}: This method processes chunks of the masked letter and passes them to the MLM model, which in turn outputs probabilities for each masked entity. The algorithm replaces each entity either deterministically (by selecting the most probable word) or stochastically (by sampling from the probability distribution). A trade-off emerges where stochastic selection enhances diversity but may slightly reduce fidelity by introducing additional noise in the generations.
    
    \item \textit{Iterative Mask Filling} \cite{kesgin2023iterative}: This method processes each masked entity iteratively within a context window. Preceding masked words are replaced with their selected counterparts, while future masked entities keep their original values until processed. By focusing on one masked entity at a time, this method provides a stronger context for the MLM model to enhance the generations' quality. Moreover, as each entity is replaced iteratively, it further motivates diversity in the output. Replacements can also be chosen deterministically or stochastically as with the previous method.
\end{itemize}

\pagebreak

\section{Experimental Setup}
\label{sec:experimental_setp}
This section outlines the dataset and training process used for the MLM model, and describes the four system instances evaluated in our experiments.

\subsection{Datasets} 
\label{sec:dataset} 
Both model training and evaluation are performed on the i2b2 2014 shared task dataset for PHI de-identification \cite{stubbs2015annotating,stubbs2015automated}, which contains 1304 English clinical records from 296 diabetic patients, including discharge summaries, admission notes, and doctor correspondences. It is pre-divided into 790 training and 514 testing samples.

This dataset offers a diverse set of clinical conditions and treatments, allowing our model to generate diverse synthetic samples. All records come with PHI annotations that are compliant with HIPAA standards. In addition, some extra PHI sub-categories  are considered and annotated to further ensure patient protection. Details on annotation categories are provided in Appendix \ref{appendix:dataset}.

\subsection{System Instances} 
\label{sec:systems} 
We evaluate four system instances with varying \textit{Masker} ratios and \textit{Mask-Filling algorithms}: System\_S\_0.5, System\_S\_0.7, System\_I\_0.7, and System\_I\_0.9. Descriptions of these configurations are provided in Appendix \ref{appendix:system_instances}. Masking ratios were chosen based on findings from \newcite{micheletti2024exploration} and can be adjusted for specific applications. 

Details on the hyperparameter tuning and training of the MLM model are given in Appendix \ref{appendix:training}.

\section{Experiments and Results} 
\label{sec:results}
We evaluate all system instances across three key aspects: resemblance to real data, data utility, and privacy. Details on each evaluation metrics are provided in Appendix \ref{appendix:eval_metrics}, and examples of generated synthetic letters are displayed in Appendix \ref{appendix:examples_letters}.

\subsection{Lexical Similarity Evaluation against References}
\label{sec:lex_sim}
The ROUGE and BERTScore metrics of the four system instances are shown in Table \ref{fig:lexical_eval}. 

Greater masking ratios result in lower ROUGE and BERTScore values due to the additional noise they convey. This confirms the trade-off between diversity and fidelity outlined in Section~\ref{sec:design}.

Moreover, instances with iterative mask filling demonstrate better robustness than ones with simultaneous filling regarding lexical similarity to real data. In fact, at the same masking ratio (0.7), the former achieves higher ROUGE scores by over 3 points and higher BERTScore by over 0.3. This highlights the advantage of iterative mask filling, where each masked token is surrounded by original or predicted tokens, enhancing context and reducing uncertainty. 
Furthermore, at a masking ratio of 0.9, iterative systems show a smaller decline in BERTScore (0.04) compared to ROUGE scores (4 points), indicating that while the generated letters are lexically further away from the original ones, their meaning is mostly preserved. 

In fact, these results are consistent with those in Appendix~\ref{app:lex_sim}, which evaluates lexical differences by comparing word overlaps between real and synthetic datasets.

In general, all instances could effectively balance their diversity with the amount of core information retained. The results demonstrate a clear trade-off between the two, which can be adjusted by tuning masking ratios and filling methods, providing flexibility for various applications.

\begin{table}[b]
\centering
\begin{tabular}{l||l|l|l|l}
               & RGE1 & RGE2 & RGE-L & BERTS \\ \hline
Sys\_S\_0.5 & 0.861   & 0.760   & 0.852   & 0.729     \\
Sys\_S\_0.7 & 0.828   & 0.703   & 0.815   & 0.674     \\ \hline
Sys\_I\_0.7 & 0.852   & 0.732   & 0.841   & 0.706     \\
Sys\_I\_0.9 & 0.826   & 0.686   & 0.811   & 0.668     \\
\end{tabular} \vspace{1em}
\caption{Lexical similarities of the generated synthetic letters against references on the testing dataset.}
\label{fig:lexical_eval}
\end{table}

\subsection{Readability Evaluation against References}
According to the results of the readability evaluation shown in Table~\ref{fig:read_eval}, synthetic letters are, on average, easier to read than the original ones. Additionally, higher masking ratios tend to improve readability, as the MLM model often replaces masked tokens with simpler, more common words.

When comparing systems against each other, no clear winner emerges. This flexibility turns out to be advantageous, as it indicates that users can tune the trade-off between diversity and fidelity without sacrificing readability.

\begin{table}[t]
\centering
\begin{tabular}{l||l|l|l}
               & FRE      & FKG   & SMOG    \\ \hline
System\_S\_0.5 & 64.024   & 7.647   & 10.823 \\
System\_S\_0.7 & 65.091   & 7.466   & 10.696 \\ \hline
System\_I\_0.7 & 63.792   & 7.707   & 10.878 \\
System\_I\_0.9 & 64.294   & 7.636   & 10.832 \\ \hline
\textbf{References} & 61.597 & 8.06 & 11.067 \\
\end{tabular} \vspace{.2em}
\caption{Readability scores of the generated synthetic letters against references on the testing dataset.}
\label{fig:read_eval}
\end{table}

\subsection{Data Utility Evaluation}
This phase evaluates how well the synthetic data capture critical characteristics of real data by comparing a medical NER model trained on synthetic data against one trained on real data.

\subsubsection{Downstream NER Task}
\label{sec:downstream}
In this downstream task, the testing set is first split into training and testing subsets. Original letters are processed through our system to generate synthetic counterparts. Both real and synthetic letters are then passed through SciSpacy\footnote{https://allenai.github.io/scispacy/} (\textit{en_ner_bc5cdr_md}), an NER model trained on the BC5CDR corpus (with an F1 score of 0.84), to detect DISEASE and CHEMICAL entities. Entities extracted from both the original and synthetic data are then used to create two datasets for training SpaCy\footnote{https://spacy.io/} models from scratch. That is, one model is trained on the entities extracted from the real data and another on the entities extracted from the synthetic data. Finally, both instances of SpaCy are evaluated on the testing subset.

To assess the impact of data augmentation, the experiment is also repeated with double the amount of synthetic letters per original letter. 

Note that, while SciSpacy's extraction errors may propagate, we expect them to be proportional across real and synthetic data.

\subsubsection{Results of Downstream Task}
Table \ref{tab:downstream_results} shows the results of the downstream task. 
All systems achieved performance comparable to models trained on real data. Interestingly, higher masking ratios improved F1 scores, which may be due to increased diversity in the generated synthetic samples, providing more diverse samples for SpaCy to train on.

Furthermore, augmenting synthetic data to twice the original amount further improved the F1 score to 0.836, which is only 0.006 lower than models trained on real data.

\pagebreak

\begin{table}[t]
\centering
\begin{tabular}{cl||l|l|l}
               && Precision& Recall  & F1    \\ \hline

                  & System\_S\_0.5 & 0.842    & 0.792   & 0.816 \\
                  & System\_S\_0.7 & 0.851    & 0.797   & 0.823 \\
\textbf{x1}       & System\_I\_0.7 & 0.831    & 0.812   & 0.821 \\
                  & System\_I\_0.9 & 0.846    & 0.810   & 0.827 \\ \hline

                  & System\_S\_0.5 & 0.844    & 0.800   & 0.821 \\
                  & System\_S\_0.7 & 0.850    & 0.805   & 0.828 \\ 
\textbf{x2}       & System\_I\_0.7 & 0.838    & 0.819   & 0.829 \\
                  & System\_I\_0.9 & 0.855    & 0.819   & \textbf{0.836} \\ \hline

& \textbf{References} & 0.86 & 0.824 & 0.842 \\
\end{tabular} \vspace{1em}
\caption{Average Precision, Recall and F1 score for two labels (DISEASE and CHEMICAL) using Synthetic data $\times1$, $\times2$ and Real data, on the testing dataset.}
\label{tab:downstream_results}
\end{table}

\subsection{Data Privacy Evaluation}
In the privacy evaluation, we first calculate the de-identification rate of our system, i.e., the accuracy of the Masker in identifying all PHI from the testing dataset. The Masker achieves a recall of 0.92 across all PHI categories (including extra sub-categories) and 0.96 for HIPAA-PHI-only categories.

Second, we evaluate the re-identification risk, i.e., the probability of the MLM model to reinsert a masked PHI. This is to ensure the privacy of the individuals whose data were used to train the system. As a result, the MLM model re-injected PHI entities of over two tokens with a rate of only 0.035. Additionally, the longest common substring analysis for PHI between original and synthetic data revealed rates as low as 0.098 (for longest common substrings of 3 tokens or more), 0.020 (for 5 tokens or more), and 0.009 (for 7 tokens or more).

These results highlight the system's effectiveness in de-identifying HIPAA-PHI entities while ensuring minimal re-identification risk.

\section{Conclusion}
\label{sec:conclusion}
In conclusion, the results demonstrated that
\textbf{(1)} the system effectively generated synthetic medical records while preserving their core medical meaning and introducing significant diversity. 
\textbf{(2)} The model's flexibility allows users to adjust the trade-off between diversity and fidelity by tuning masking ratios and mask-filling techniques, without compromising readability.
\textbf{(3)} Furthermore, the downstream evaluation showcased the system's ability to train SpaCy on a medical NER task, achieving performance comparable to models trained on real data. This underscores the quality of the synthetic records and their viability as an alternative to real data.
\textbf{(4)} Finally, the system demonstrated high effectiveness in de-identifying HIPAA-PHI entities with a recall of 0.96, while maintaining a low re-identification risk of 0.035.

\subsection{Limitations and Future Work}
Upon careful analysis of the generated samples, we observed challenges in consistently filling temporal information and aligning it with the original data. Additionally, maintaining coherence in interconnected events, such as accurately assigning two names within a discussion, is sometimes problematic when relevant context is not available within the generation window.
Future improvements could involve integrating a logic-based component to fill in temporal information, further reducing re-identification risk and ensuring temporal consistency. Another potential enhancement is passing the type of entity to be replaced to the MLM model, which may improve the accuracy of PHI replacements and overall generation quality.

Regarding the MLM model, future work could explore using large language models to process masked letters through guided prompt instructions. This approach would focus on the mask-filling task, enabling a more comprehensive comparison of CLMs and MLMs at generating synthetic data with controlled fidelity and diversity. In this scenario, the Masker would remain unchanged while the MLM model would be replaced with a CLM and the Mask-filling algorithm with an instruction prompt.

Finally, note that the results may not be fully generalisable, as a single dataset was used due to computational constraints. Expanding the evaluation to a broader range of downstream tasks and datasets would provide a more comprehensive understanding of the system's potential applications. For instance, future works could apply the system to specialised datasets, such as radiology or oncology. This would require to change for appropriate NER models (e.g., \textit{Stanza Radiology} or \textit{Stanza Bionlp13cg}) in order to extract relevant medical information. However, this may involve exploring new masking ratios for both the medical NER model and the POS tagger to refine performance.

\clearpage
\bibliography{main}

\clearpage
\appendix
\section{Annotation Categories in Dataset}
\label{appendix:dataset}
As explained in section \ref{sec:dataset}, the provided annotations are HIPAA-PHI compliant and include additional sub-categories to further ensure the patients’ protection. Below are listed all categories of annotations:

NAME (types: PATIENT, DOCTOR, USERNAME); PROFESSION; LOCATION (types: ROOM, DEPARTMENT, HOSPITAL, ORGANIZATION, STREET, CITY, STATE, COUNTRY, ZIP, OTHER); AGE; DATE; CONTACT (types: PHONE, FAX, EMAIL, URL, IPADDRESS); IDs (types: SOCIAL SECURITY NUMBER, MEDICAL RECORD NUMBER, HEALTH PLAN NUMBER, ACCOUNT NUMBER, LICENSE NUMBER, VEHICLE ID, DEVICE ID, BIOMETRIC ID, ID NUMBER).

Out of these categories, only the following correspond to the HIPAA-PHI categories: NAME-PATIENT, LOCATION-STREET, LOCATION-CITY, LOCATION-ZIP, LOCATION-ORGANIZATION, AGE, DATE, CONTACT-PHONE, CONTACT-FAX, CONTACT-EMAIL, as well as all ID sub-categories.

\section{Details on Hyperparameter tuning and Training}
\label{appendix:training}
During training, we perform a grid search to select the most optimal set of hyperparameters from the following values: $\alpha \in \{1\times10^{-4}, 5\times10^{-5}, 3\times10^{-5}\}$, $\beta \in \{8, 16\}$, $\phi \in \{0.75, 1.0\}$ and $\psi \in \{0.30, 0.50\}$; where $\alpha$ is the learning rate of the MLM model, $\beta$ is the training batch size, $\phi$ is the PHI's masking proportion and $\psi$ is the overall masking probability. For convenience, we select the optimal number of training epochs through early stoppage with a patience of $p=2$. While we agree that more advanced hyperparameter search methods exist, such as Bayesian Optimisation \cite{turner2021bayesian} or Optuna \cite{akiba2019optuna}, we decided to opt for grid search due to computational limitations.

We split the dataset into 80\% training and 20\% validation, using a random split. We once again recognise that k-fold cross-validation is more accurate, but are constrained by the same computational resources. For each possible set of hyperparameters, a new instance of the system is created. Then, during its training, training samples are re-processed at each epoch with a random masking of up to $\psi$ percent, including $\phi$ percent of all PHI entities. This allows the model to see varied versions of the same sample, increasing the diversity of cases it can learn from and reducing overfitting. In contrast, the validation set is masked consistently across all epochs to ensure fair comparison.

We evaluate each instance using perplexity as it reflects the MLM model's confidence. Once the best hyperparameters are identified, we merge the training and validation sets and retrain the best model on the full dataset.

\section{Details on System Instances used throughout Experiments}
\label{appendix:system_instances}
Below are described the four distinct system instances presented in section \ref{sec:systems}.

\begin{itemize}
    \item \textbf{System\_S\_0.5}: This instance masks all PHI entities and none of the medical entities captured by the NER. However, it masks 50\% of NOUNS, VERBS and ADJECTIVES for moderate diversity. In addition, it uses the Simultaneous Chunk Filling algorithm for mask-filling with stochastic selection to increase diversity.
    \item \textbf{System\_S\_0.7}: Similarly to \textit{System\_S\_0.5}, this instance masks all PHI entities and none of the medical entities captured by the NER. However, it masks 70\% of NOUNS, VERBS and ADJECTIVES for increased diversity, and uses the same Simultaneous Chunk Filling algorithm for mask-filling with stochastic selection to increase diversity.

    \item \textbf{System\_I\_0.7}: This instance masks all PHI entities and none of the medical entities captured by the NER. It masks 70\% of NOUNS, VERBS and ADJECTIVES, and uses Iterative Mask Filling with stochastic selection to increase diversity.
    \item \textbf{System\_I\_0.9}: Similarly to \textit{System\_I\_0.7}, this instance masks all PHI entities and none of the medical entities captured by the NER. However, it masks 90\% of NOUNS, VERBS and ADJECTIVES, and uses the same Iterative Mask-filling technique with stochastic selection to increase diversity.
\end{itemize}

\section{Description of Evaluation Metrics}
\label{appendix:eval_metrics}
We describe below the three aspects on which our evaluation is based, namely resemblance/similarity to real data, data utility, and privacy.

\textbf{Lexical similarity to reference} evaluates the ability of our synthetic data to resemble the statistical characteristics of real data at both variable and record levels. This includes lexical similarities such as "how much information is retained from the original data?", "how much overall meaning is maintained post-synthetisation?" and "how much diversification and deviation (prevalence) was generated?", which are evaluated with ROUGE, BERTScore and ROUGE metrics, respectively. It further includes readability comparisons such as "how easily can the text be read?" and "what academic level do you need to read the document?", which are evaluated with FRE\footnote{Flesh Reading Ease} and the pair FKG\footnote{Flesch-Kincaid Grade}-SMOG, respectively.

\textbf{Data utility} measures how well the generated data captures the critical characteristics present in the real data. To assess this characteristic, we evaluate the extent to which our synthetic records retain the capability of training machine learning models that perform comparably to those trained using real data. This is done through a downstream NER task, similarly to \newcite{belkadi2023generating,micheletti2024exploration}. 

\textbf{Data privacy} evaluation is crucial when considering the sharing of synthetic medical data. As our current dataset has been labelled by multiple professionals following the official HIPAA-PHI de-identification rules, we evaluate the privacy level of our model by calculating the F1 score to how much of the PHIs were identified and replaced by our system according to the annotated data, and how much re-identification occurred on average.

\section{More Lexical Similarity Results}
\label{app:lex_sim}
Below are additional results on lexical similarities.

\begin{table}[h]
\centering
\begin{tabular}{l||l|l|l|l}
               & Top 5   & Top 20  & Top 50 & Top 100  \\ \hline
System\_S\_0.5 & 3.848   & 15.593  & 38.420 & 78.670   \\
System\_S\_0.7 & 3.601   & 14.607  & 35.971 & 73.695   \\
System\_I\_0.7 & 3.712   & 15.095  & 37.233 & 76.093   \\
System\_I\_0.9 & 3.537   & 14.551  & 35.510 & 72.298   \\
\end{tabular} \vspace{1em}
\caption{Average number of overlap between the top 5, 20, 50 and 100 words identified across the real and synthetic datasets, without stopwords.}
\label{fig:overlap_results}
\end{table}

\pagebreak
\section{Examples of generated synthetic letters}
\label{appendix:examples_letters}

\begin{figure*}[p]
    \centering
    \begin{minipage}{0.47\textwidth}
        \centering
        \includegraphics[width=\textwidth]{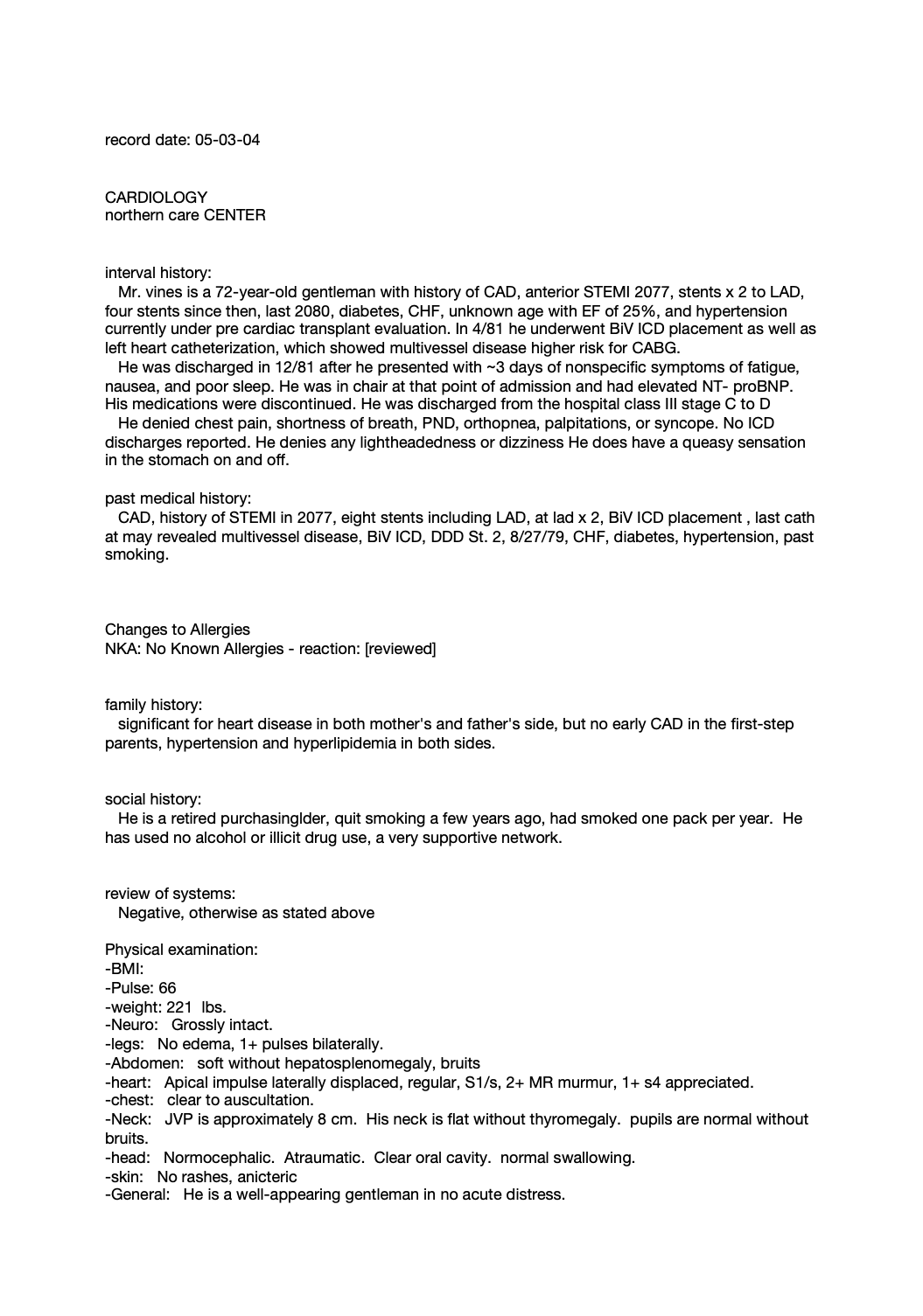} 
    \end{minipage}\hfill
    \begin{minipage}{0.47\textwidth}
        \centering
        \includegraphics[width=\textwidth]{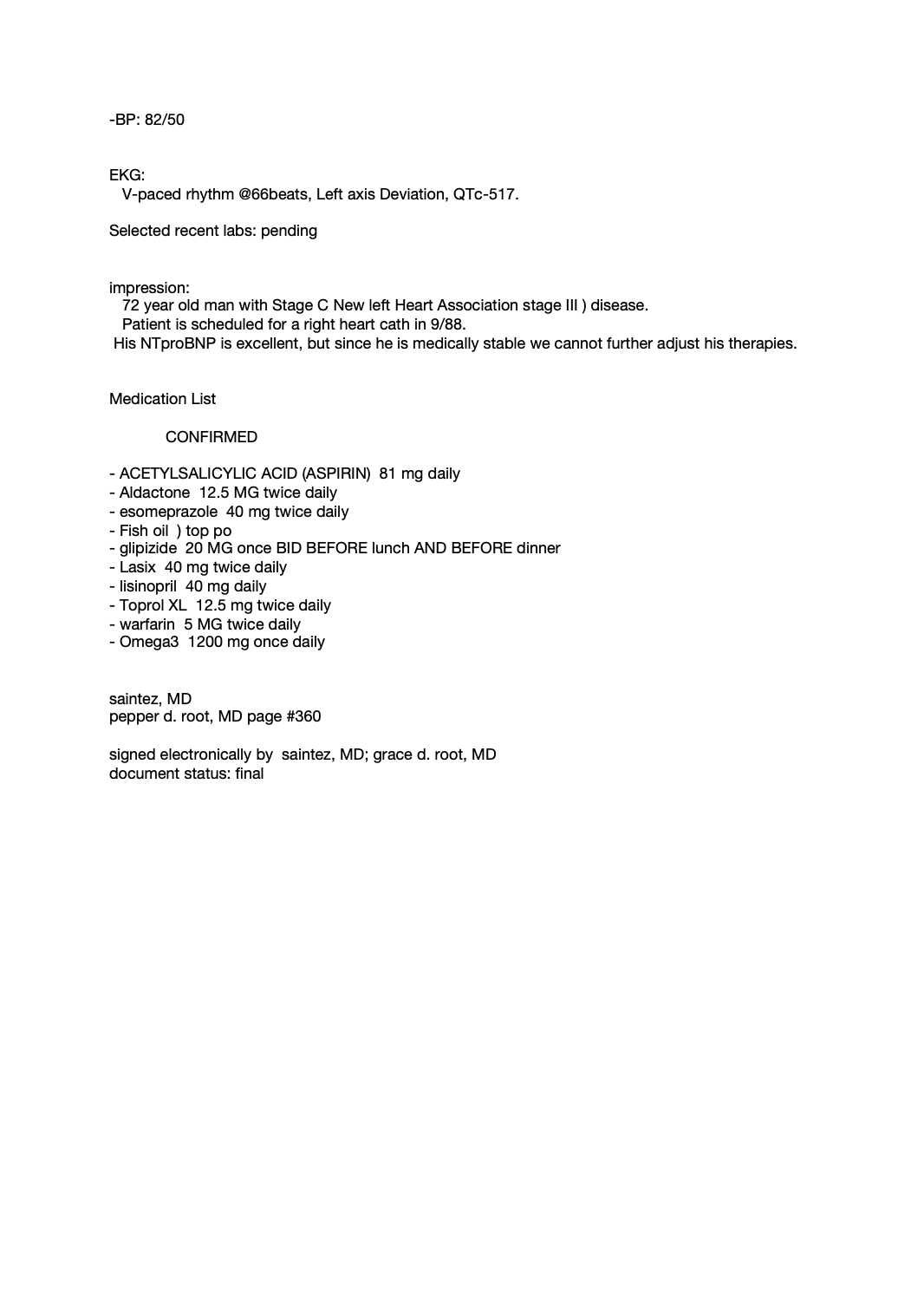}
    \end{minipage}
    
    \begin{minipage}{0.47\textwidth}
        \centering
        \includegraphics[width=\textwidth]{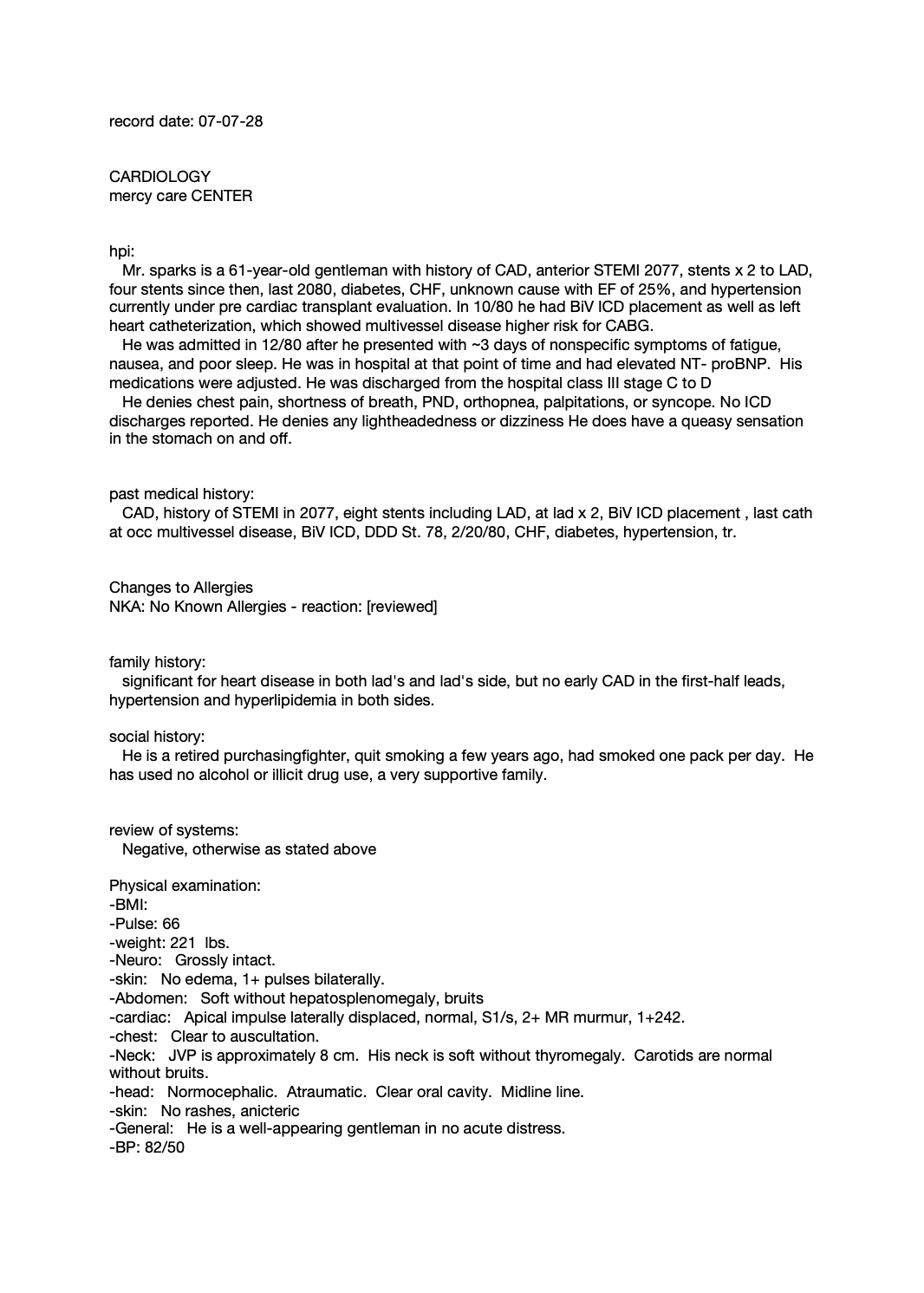} 
    \end{minipage}\hfill
    \begin{minipage}{0.47\textwidth}
        \centering
        \includegraphics[width=\textwidth]{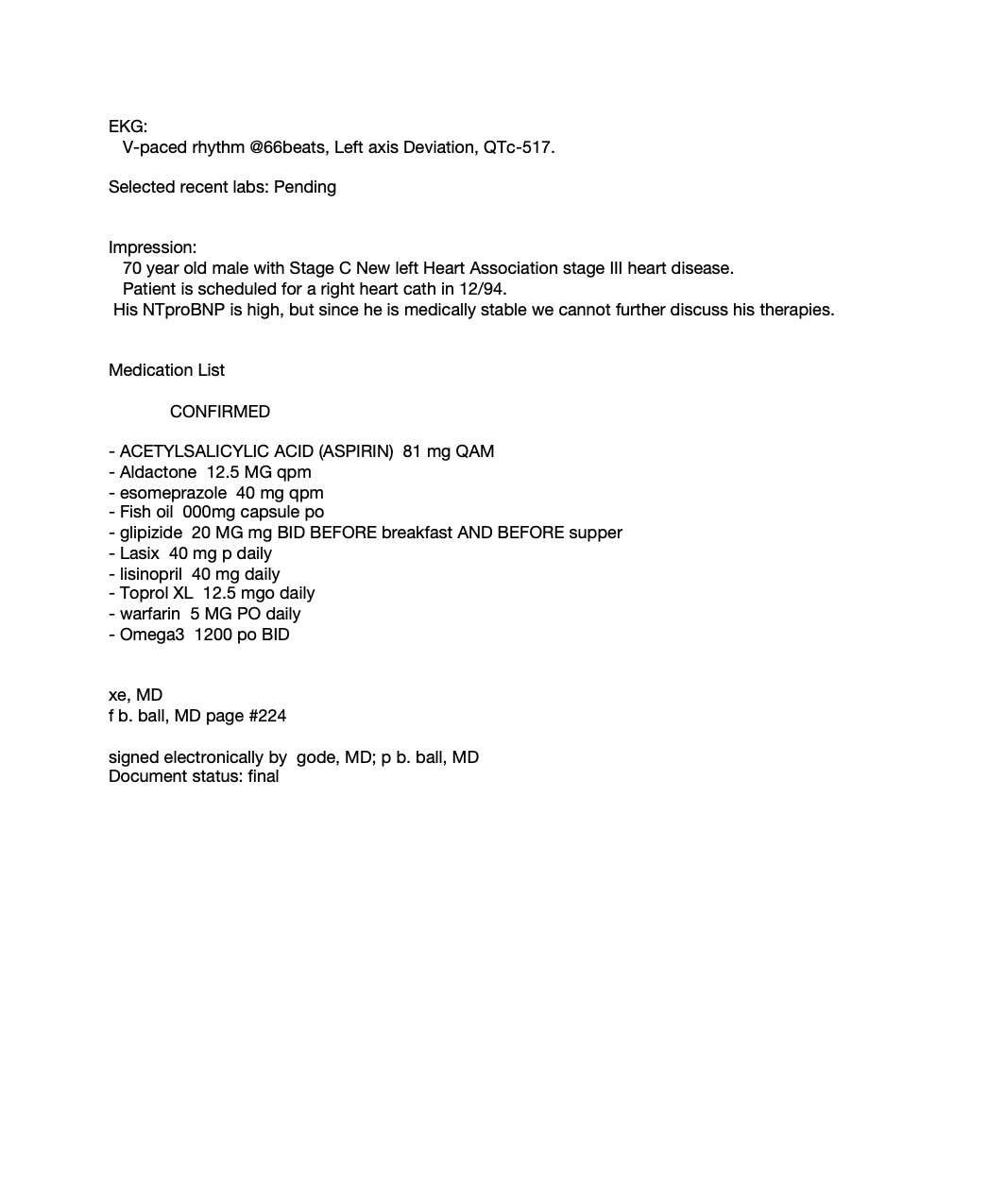}
    \end{minipage}

    \caption{Synthetic letters generated from letter 201-03 using System\_I\_0.7 \textbf{(top)} and System\_S\_0.5 \textbf{(bottom)}.}
\end{figure*}

\end{document}